\documentclass[runningheads]{llncs}
\usepackage{graphicx}
\usepackage{amssymb,amsmath,array}
\usepackage{color}
\usepackage{inputenc}
\begin{document}
\title{Mise en abyme with artificial intelligence: \\ how to predict the accuracy of NN, \\ applied to hyper-parameter tuning}
\titlerunning{How to predict the accuracy of NN}
\author{Giorgia Franchini\inst{1} \and
Mathilde Galinier\inst{1,2} \and
Micaela Verucchi\inst{1}}
\authorrunning{G. Franchini, M. Galinier, M. Verucchi}
\institute{Universit\`{a} degli studi di Modena e Reggio Emilia - PhD Student \\
Via Universit\`{a}, 4, 41121 Modena - Italy \and
Marie Sklodowska-Curie fellow of the Istituto Nazionale di Alta Matematica}
\maketitle              % typeset the header of the contribution

\begin{abstract}

%In the context of deep learning, the costliest phase from a computational point of view is the full training of the learning algorithm. However, this process is to be used a significant number of times during the design of a new artificial neural network, leading therefore to extremely expensive operations. Here, we propose a low-cost strategy to predict the accuracy of the algorithm, based only on its initial behaviour. To do so, we use both curve fitting and Support Vector Machines techniques, the latter being trained on a beforehand created database. We applied this approach to the hyper-parameter optimisation of a convolutional neural network for the classification of the databases MNIST and CIFAR-10, and found the hyper-parameter settings corresponding to the optimal accuracies, already known in literature. \\

In the context of deep learning, the costliest phase from a computational point of view is the full training of the learning algorithm. However, this process is to be used a significant number of times during the design of a new artificial neural network, leading therefore to extremely expensive operations. Here, we propose a low-cost strategy to predict the accuracy of the algorithm, based only on its initial behaviour. To do so, we train the network of interest up to convergence several times, modifying its characteristics at each training. The initial and final accuracies observed during this beforehand process are stored in a database. We then make use of both curve fitting and Support Vector Machines techniques, the latter being trained on the created database, to predict the accuracy of the network, given its accuracy on the primary iterations of its learning. This approach can be of particular interest when the space of the characteristics of the network is notably large or when its full training is highly time-consuming. The results we obtained are promising and encouraged us to apply this strategy to a topical issue: hyper-parameter optimisation (HO). In particular, we focused on the HO of a convolutional neural network for the classification of the databases MNIST and CIFAR-10. By using our method of prediction, and an algorithm implemented by us for a probabilistic exploration of the hyper-parameter space, we were able to find the hyper-parameter settings corresponding to the optimal accuracies already known in literature, at a quite low-cost. \\


\end{abstract}

\section{Introduction}

During the last decades, machine learning algorithms and deep neural networks have shown remarkable potentials in numerous fields. However, despite their success, algorithms of this kind may still be hard to design, 
and their performance usually highly depend upon the choice of numerous criteria (learning rate, optimiser, structure of the layers, etc.), called hyper-parameters. In practice, finding an optimal combination of those hyper-parameters can often make the difference
between bad or average results and state-of-the-art performance. 
The search of this optimal setting can be performed by maximising the accuracy $f$ of the given learning algorithm. However, this process is made particularly arduous in that the maximisation of such a function is usually very expensive. 

On the basis of that observation, we decided to develop a strategy for predicting the value of the objective function $f$ at low-cost, in an off-line fashion. Based on Support Vector Machine (SVM) and curve fitting, this method enables to obtain a prediction of the accuracy of an artificial neural network (NN), exploiting only the behaviour of the NN over the first epochs of its learning. To the best of our knowledge, no other article in the literature mentions approaches for a low-cost approximation of the accuracy of a NN, and in this respect, the ideas lying behind our method may constitute a breakthrough in the domain of artificial intelligence. The algorithm we propose can be of particular interest for a rather fast quality assessment of a new learning algorithm. Specially, it may facilitate hyper-parameters optimisation in an interesting way.

The Section 2 of this article describes the main steps of our method for predicting the accuracy of a given NN. Section 3 deals with an application of our method to a topical issue : hyper-parameter optimisation. Section 4 shows the results of the experiments on the MNIST and CIFAR-10 databases. Finally, the paper ends with concluding remarks in Section 5.

\section{Methodology}\label{our_method}

%\subsection*{Creation of the database} \label{creation_database}
 
A database is created beforehand, by training the NN algorithm of interest up to convergence, with different sets of randomly chosen hyper-parameters. The prediction accuracies of the NN algorithm are gathered throughout this process, and eventually constitute a database on which the hereinafter-described method is applied. Please note that a new database is to be created every time a new set of training/test samples is considered.  
%on which a SVM algorithm is trained.

\subsection*{Between SVM and curve fitting}

The goal is to be able to predict the final accuracy of the NN algorithm after convergence, based only on its behaviour over the first epochs of its learning.
To do so, we have combined two techniques, well-known in the literature: Support Vector Machines (SVM) \cite{Vapnik} and curve fitting \cite{Arlinghaus}.

We first present a theoretical background of the mentioned techniques, and then indicate how we used them in this work. \\

%a support vector machine [3] Intuitively, a good separation is achieved by the hyperplane that has the largest distance to the nearest training-data point of any class (so-called functional margin)

A SVM is a discriminative classifier formally defined by a separating hyper-plane. In other words, given labelled training data (supervised learning), the algorithm outputs an optimal hyper-plane which categorises new examples. SVM algorithms are characterised by the usage of kernels, the sparseness of the solution and the capacity control obtained by acting on a margin, or on the number of support vectors. SVM can be applied not only to classification problems but also to the case of regression. 
%Still, they contain all the main features that characterise maximum margin algorithms: a non-linear function is learned by linear learning machine mapping into high dimensional kernel induced feature space. The capacity of the system is controlled by parameters that do not depend on the dimensionality of feature space. 
One of the most important ideas in SVM for regression is that presenting the solution by means of small subsets of training points gives enormous computational advantages. 
%Using the $\epsilon$-intensive loss function, we ensure the existence of a global minimum, 
%and the {\color{red}optimisation of a reliable generalisation bound}, 
%where the $\epsilon$ parameter controls the width of the $\epsilon$-insensitive zone, used to fit the training data. 

We consider the well-known dual problem of the SVM algorithm \cite{Vapnik}, whose solution is given by:
\begin{equation*}
    f(\mathbf{x})=\sum_{i=1}^{n_{SV}}(\alpha_i-\alpha_i^*)K(\mathbf{x_i},\mathbf{x}) + b
\end{equation*}
where $n_{SV}$ is the number of support vectors, $\alpha$ and $\alpha^\ast \in \mathbb{R}^{n_{SV}}$ are the multipliers of the support vectors and $b\in\mathbb{R}$ is a constant. $K(\mathbf{x_i},\mathbf{x})$ is the kernel function, that we can choose to be linear, polynomial or Gaussian. In particular the Gaussian kernel form is
\begin{equation*}
   K(\mathbf{x_i},\mathbf{x})=\exp({-\gamma\lVert \mathbf{x_i}-\mathbf{x} \rVert^2}), \text{ where } \gamma \text{ is a free positive parameter.} 
\end{equation*}

Curve fitting is the process of constructing a curve that has the best fit to a series of data points, possibly subject to constraints. In other words, the goal of curve fitting is to find the appropriate parameters that, based on a chosen model, describe experimental data as precisely as possible. The values of such parameters are usually computed by the Least Squares (LS) method, which minimises the square of the error between the original data and the values predicted by the model. 
In this work, the fitting function is chosen to have the form :
\begin{equation}
    g(x) = \alpha x^\beta
    \label{fittingfun}
\end{equation}
where $\alpha$ and $\beta$ are the parameters to be computed by non-linear LS. \\

Our methodology is based on the following procedure: knowing the accuracies over the first epochs of its training process, the accuracy reached by an NN after convergence is predicted thanks to an SVM algorithm, beforehand trained on the created database. This value is considered as an approximation of the objective function $f$ of interest, in the case it is neither greater than $1$, nor smaller than the maximum of the initial observed accuracies. Otherwise, the final accuracy of the NN is predicted based on a curve fitting strategy: given the values of the accuracies of the first epochs,
the corresponding curve is fitted thanks to the function (\ref{fittingfun}). Once the parameters $\alpha$ and $\beta$ are computed, the function $g$ is used to predict the value of the accuracy after convergence of the algorithm. It is to be noted that the parameters are subject to the following constraints: 
\begin{equation}
\left \{
\begin{array}{c c}
    \frac{acc\_max}{fin\_epoch} < \alpha  \\
    0 < \beta < 1
\end{array}
\right.
\end{equation}
where $acc\_max$ is the maximum among the accuracies over the first epochs, and $fin\_epoch$ is the maximum number of epochs that are allowed to be seen by the learning algorithm. In particular, the first constraint forces the curve to reach an accuracy at the final epoch higher than the one already observed during the first epochs.

Combining those two different techniques for the prediction of the final accuracy makes the prediction process more efficient, this is why we decided not to use SVM only. Furthermore, the curve fitting method seemed to us appealing insofar as the fitting function can easily be constrained, as explained above. This heuristic choice was then validated by the experiments. 

This method can be convenient from a computational point of view because, once the database is created and the SVM is trained, it allows to have new evaluations of the objective function in a short time. The time taken in creating the database must be evaluated; and obviously, the convenience of such a method increases as the complexity of the problem increases.

\section{Application : from prediction to hyper-parameter optimisation}

Being able to predict the accuracy of an NN for a given hyper-parameter setting is particularly advantageous for finding the optimal combination of hyper-parameters. In this section, we aim to apply the hereinabove described method to the hyper-parameter optimisation (HO) of a Convolutional Neural Network (CNN). 
We first describe the state-of-the-art HO approaches. Secondly, we present the algorithm we designed to find the optimised hyper-parameters.

\subsection{Related works}
Two of the most intuitive and widely spread methods are the grid search and the random search \cite{Bergstra_2012}. These techniques however are not well-suited in applications where a given set of hyper-parameters is costly to evaluate.
As a result, Sequential Model-Based Optimisation (SMBO) \cite{Hutter_2011} algorithms have been employed in many settings when the performance evaluation of a model is expensive. They approximate the black-box objective function $f$ that is to be maximised by a surrogate function, cheaper to evaluate.
At each iteration of the algorithm, the new point where the surrogate is to be evaluated is chosen by maximising a chosen criterion.
Several SMBO algorithms have been proposed in the literature, and differ in the criteria by which they optimise the surrogate, and in the way they model the surrogate given the observation history. Two of the most famous SMBO approaches are the Bayesian Optimisation approach \cite{Shahriari_2016,Mockus_1978} and the Tree-structured Parzen Estimator strategy \cite{Begstra_2011}.

More recently, new hyper-parameter optimisation methods based on reinforcement learning have emerged \cite{Zoph_2017,Baker_2017,Zhong_2018,Cai_2108}. The goal for most of them was to find the neural network (NN) or CNN architectures that are likely to yield to an optimised performance. Thus, they were seeking the appropriate architectural hyper-parameters, as the number of layers or the structure of each convolutional layer, but many other hyper-parameters
such as the learning rate and regularisation parameters are manually
chosen in the end. 
In any case, while all the above-mentioned strategies aim to evaluate the expensive objective function $f$ (which, in the case of a NN or CNN is the prediction accuracy) as seldom as possible, to the best of our knowledge very few algorithms offer a method to reduce the evaluation cost of $f$.

\subsection{Our method applied to HO}
Here, we assume that we are given a CNN, and we seek its optimal hyper-parameters. Basing the optimisation of these hyper-parameters on our above-mentioned method actually implies the choice of other parameters, such as the type of kernel function, or the form of the fitting function (\ref{fittingfun}). However, it is to be observed that those new parameters can be much more easily chosen. For instance, SVM for regression is yet a very well known method in literature and offers a good robustness in its results with respect to its hyper-parameters. On the contrary, this is definitely not the case for NN in general, which makes the choice of their parameters much harder.

We focus on the optimisation of the learning rate, the optimiser and the mini-batch size leading to the best accuracy, but this method could be extended to other hyper-parameters (layer number, number of neurons in each layer, activation function).

The procedure we propose consists in the following steps. First, a database is created, as above-mentioned, and an SVM algorithm is trained on it. We chose to create this database using 10$\%$ of the possible settings of hyper-parameters, randomly selected with uniform probability.
Although time-consuming, this preliminary step may lead to a less expensive hyper-parameter optimisation process than the ones already known in the literature. 

In our HO approach, each hyper-parameter $i$ of interest is represented as a vector $V_i$ containing values in a range chosen by the experimenter. To this vector corresponds a second vector of probabilities $P_i$, initialised with uniform distribution.
At each iteration of the exploration process, a set of hyper-parameters is randomly chosen, based on the probabilities in the corresponding vectors $P_i$. The CNN is parameterised with the selected hyper-parameters, and trained on a few epochs. Based on its behaviour at the beginning of its learning, and thanks to the methodology detailed in Section \ref{our_method}, the accuracy of the CNN after convergence of its learning is predicted, this value being considered as a reward. If the reward $r^{(t)}$ at the iteration $t$ is higher than the reward $r^{(t-1)}$ at the previous iteration, the probabilities in $P$ corresponding to the selected hyper-parameters and their neighbourhood in the vector are increased, while the other ones are penalised. On the contrary, if $r^{(t)}<r^{(t-1)}$ the probabilities of the selected hyper-parameters are penalised and the other ones are increased.
Thus, the hyper-parameters are weighted by probabilities that are modified throughout the exploration process, until one value for each hyper-parameters reaches a probability greater than some threshold $t_i$. Then, our exploration algorithm is considered to have converged.

Finally, the NN is parameterised with the sets of hyper-parameters corresponding to the 10 highest predicted final accuracies, and brought to convergence. The setting leading to the best observed final accuracy is defined as the optimal hyper-parameter setting.

\section{Experiments}

In order to evaluate and tune our method, we performed different tests.
We decided to consider a spreadly used NN, namely CNN for classification. We picked two well-known datasets, MNIST\footnote{http://yann.lecun.com/exdb/mnist/} and CIFAR-10\footnote{https://www.cs.toronto.edu/~kriz/cifar.html}, and we designed two different networks, capable of classifying images in one of 10 classes.
For the MNIST dataset, the network is composed of an input layer, two sequences of convolutional and max-pooling layers, a fully connected layer and an output layer. We make use of Rectified Linear Unit (ReLU) activations and dropout technique.
On the other hand, for the CIFAR dataset, the CNN is composed of an input layer, four sequences of convolutional and max-pooling layers, four fully connected layers and an output layer, exploiting ReLU activations, batch normalisation, and dropout. 
For the training process of the two networks, we chose to insert a form of regularisation: the early stopping. This technique is used to avoid overfitting when training a learner with an iterative method, such as gradient descent. In particular, if the training of the network is not providing better results after a determined amount of epochs, the learning process is stopped because it is considered to be stuck in a minimum and may lead to overfitting.
As already mentioned, we first needed a database to train our SVM on. 
With the underlying idea of this paper, we won a project that granted us the use of CINECA resources on the Marconi cluster\footnote{https://www.cineca.it/en/content/marconi}. This grant enabled us to perform all our tests and generate the databases for MNIST and CIFAR. The database consists in a table in which each row corresponds to a full training of a network. We picked learning rate, optimiser and batch size as hyper-parameters and we stored them along with the accuracies observed at every epoch to create the database. Actually, the number of fully trained network needed was quite small. In our case, we needed only 44 examples: we used 35 to train the SVM and 9 to test it.
Once we collected the database, we were able to understand which SVM configuration was more suitable for our problem. We tried with three different kernels: linear, polynomial and Gaussian, and we picked the last one because it outperformed the other two in terms of loss (Mean Squared Error - MSE). 

\begin{table}[h]
\begin{center}
\begin{tabular}{ c c c c c }
& Ground-truth & Linear & Polynomial & Gaussian \\
\hline
0 & 0.7129 & 0.816350 & 0.854324 & 0.713435 \\
1 & 0.1871 & 0.197152 & 0.207737 & 0.175775 \\
2 & 0.1000 & 0.200523 & 0.204122 & 0.200774 \\
3 & 0.6820 & 0.704606 & 0.599408 & 0.781832 \\
4 & 0.4340 & 0.381075 & 0.301048 & 0.456969 \\
5 & 0.6369 & 0.572801 & 0.489510 & 0.638199 \\
6 & 0.7315 & 0.747803 & 0.682245 & 0.805067 \\
7 & 0.1000 & 0.190538 & 0.203211 & 0.175411 \\
8 & 0.4783 & 0.410684 & 0.325788 & 0.491291 \\
\hline
MSE & 0 & 0.04136 & 0.1138 & 0.0320 \\
\end{tabular}
\end{center}
\caption{%Predictions on random test values of the three different SVM kernels: linear, polynomial and Gaussian.
Given the accuracies of the NN over its first three epochs, the final accuracies are predicted using the aforementioned method.  Three different SVM kernels are investigated : linear, polynomial and gaussian. The nine first lines show the predicted final accuracies of nine different test cases, the last line indicates the MSE losses corresponding to each method.}
\label{tab: svm_kernels}
\end{table}

%\begin{table}[h]
%\begin{center}
%\begin{tabular}{ c c }
%Kernel & Loss (MSE) \\
%\hline
%Linear & 0.04136 \\
%Polynomial & 0.1138 \\
%Gaussian & 0.0320 \\
%\end{tabular}
%\end{center}
%\caption{MSE losses of the three SVM kernels.}
%\label{tab: svm_losses}
%\end{table}

In Table \ref{tab: svm_kernels} the predictions of the three different kernels are reported. In this case, predictions are made on the test set and are based on three epochs, for the CIFAR dataset, the MSE losses are reported in Table \ref{tab: svm_kernels}, where is shown that the Gaussian one is the kernel that dominates the other two.

\begin{figure}[h!]
    \centering
    \includegraphics[width=3in,trim=3cm 9cm 3cm 9cm, clip]{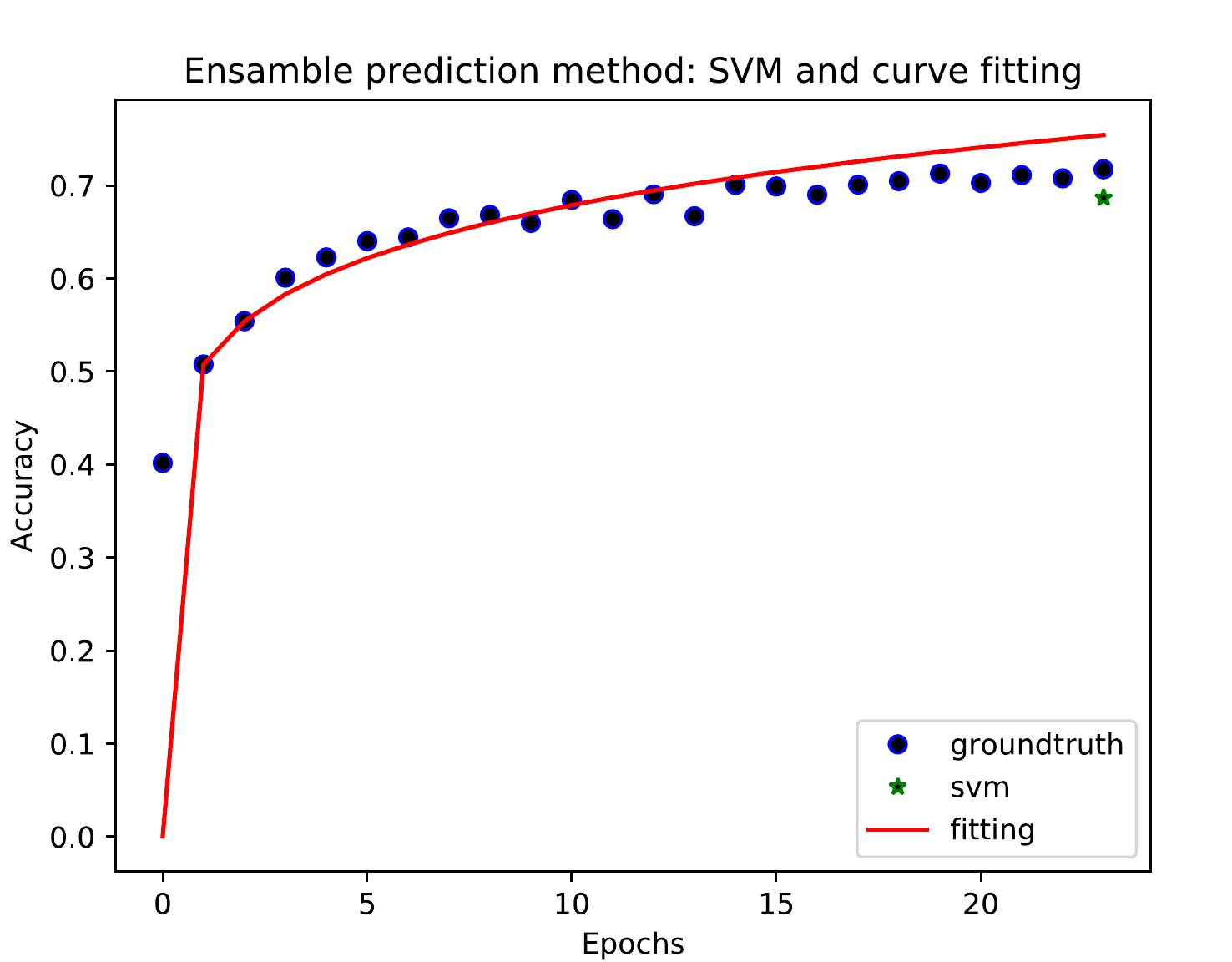}
    \caption{Prediction of curve fitting and SVM with three epoches, compared with the real accuracies.}
    \label{fig:svm_fit}
\end{figure}

We also made some tests to understand which were the most important features to feed the SVM predictor with and finally chose, given the results, to feed the method only with the epoch accuracies. Regarding the curve fitting, we also tried several functions in order to find the one that produces the best fitting (\ref{fittingfun}). Figure \ref{fig:svm_fit} shows a graphical example of the prediction made by the SVM and the curve fitting. In the chart, the full dots represent the real values of the accuracies, epoch by epoch; the star at the final epoch is the SVM prediction, while the line is curve obtained with the fitting. For both SVM and curve fitting, only the first three epochs accuracies are used to predict the accuracy at the final epoch.

\begin{figure}[h!]
    \centering
            \includegraphics[scale=0.5,width=4in,trim=2cm 4.5cm 2cm 4.5cm, clip]{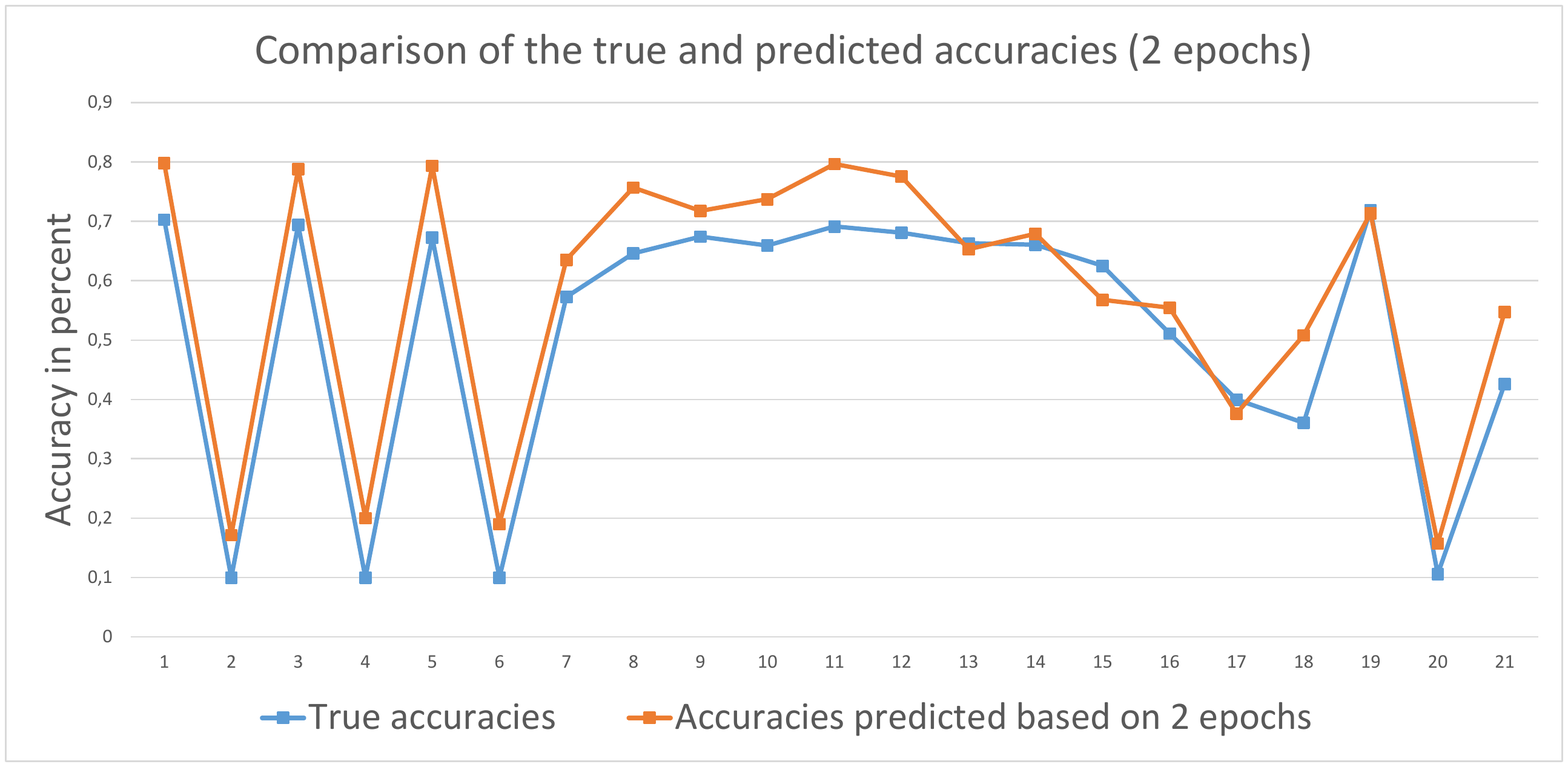}
            \includegraphics[scale=0.5,width=4in,trim=2cm 4.5cm 2cm 4.5cm, clip]{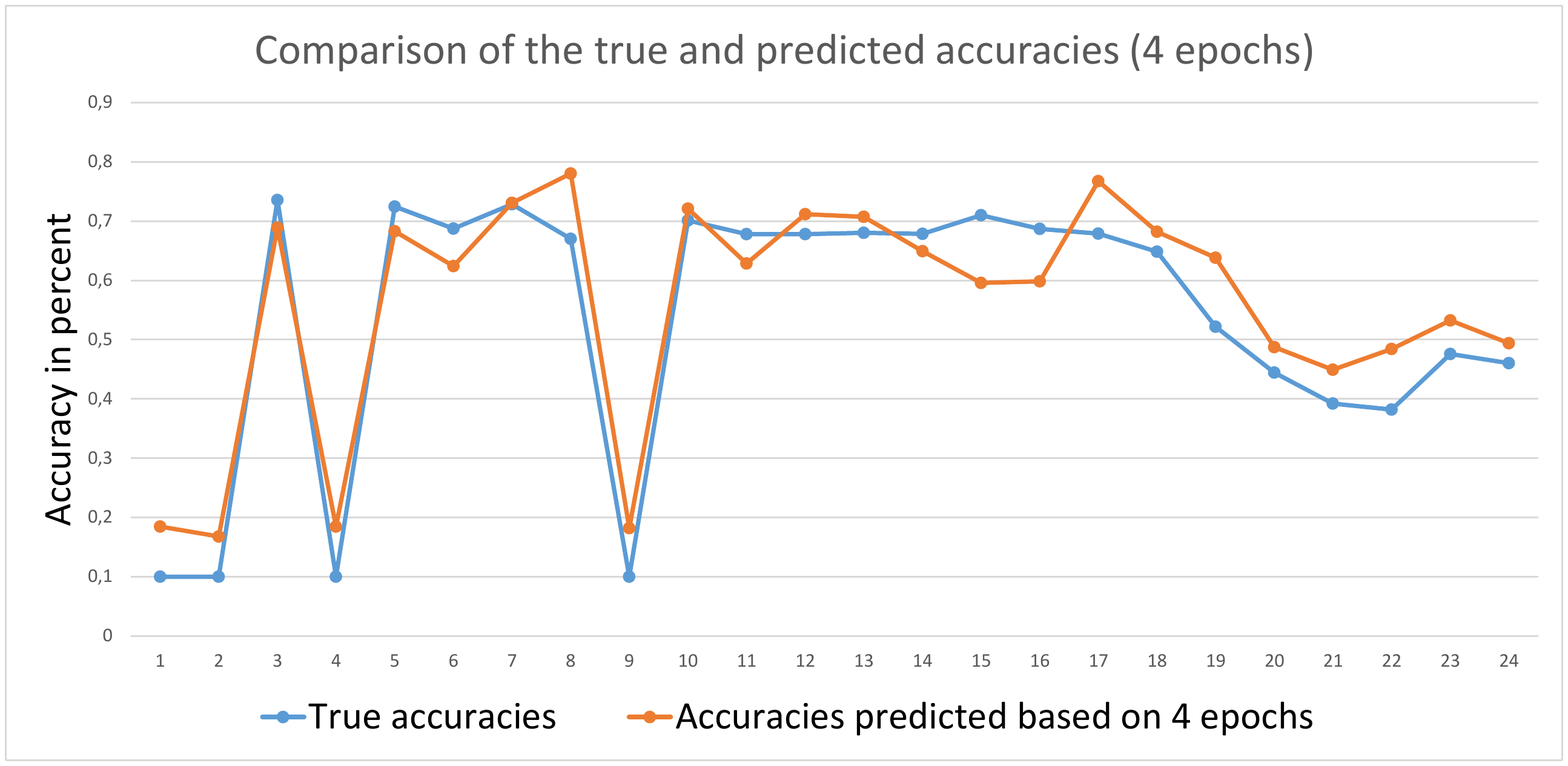}
    \caption{Comparison of the predicted final accuracy with SVM and curve fitting and the ground truth, feeding the method with only two training epochs (on the left) and four (on the right).}
    \label{fig:cifar2_4}
\end{figure}

Figure \ref{fig:cifar2_4} shows the results of our method applied to the prediction of the final accuracy of the CNN for the CIFAR-10 dataset. In this case, we use only the accuracies of the first two or four epochs of training to predict the final one. The orange line represents the prediction made by our method, the blue one the ground truth ( \textit{i.e. }the network is fully trained up to convergence with the same hyper-parameters). As it can be seen, the method can effectively provide a satisfactory prediction of the final behaviour of the network after only few epochs. 
Encouraged by those results, we applied our strategy to a challenging topical case: to tune automatically the hyper-parameters of a network, with the procedure explained in section 3. 
We performed this further test using the two first epochs to predict the final one. After 200 iterations, we were able to find the parameters that lead to the best final accuracy, confirmed by a real full convergence. The hyper-parameters provided by our method are: the learning rate is equal to 0.0008425, the mini-batch size is 128, and the chosen optimiser is ADAM.

The source code can be found at \small{https://git.hipert.unimore.it/mverucchi/optics.}

\section{Conclusion}

%On est les meilleures, des coeurs sur nous. $<3$ 

We proposed and implemented a novel approach to predict the final behaviour of a learning process. This new method exploits both SVM and curve fitting to foresee the resulting accuracy of a long method, using only some initial steps. We applied this technique to a CNN, in order to quickly understand if the training of the network will end up in a good or bad manner. The results show that the predictions achieved with our technique are quite similar to the ground-truth, and confirm that this strategy can be of particular interest in the hyper-parameter optimisation domain. Further, we will focus on a more complete procedure to automatically tune hyper-parameters, such as the number of layers or the activation function, of a network exploiting our SVM-curve-fitting predictor. 

\small{
\section*{Acknowledgements}
The research leading to these results has received funding from the European Union's Horizon 2020 Programme under the CLASS Project (https://class-project.eu/), grant agreement n $780622$.\\
This work was also partially supported by INdAM-GNCS (Research Projects 2018).
Furthermore, it was partially supported by INdAM Doctoral Programme in Mathematics and/or Applications Cofunded by Marie Sklodowska-Curie Actions (INdAM-DP-COFUND-2015) whose grant number is $713485$.
}

% ****************************************************************************
% BIBLIOGRAPHY AREA
% ****************************************************************************

\begin{footnotesize}

% IF YOU DO NOT USE BIBTEX, USE THE FOLLOWING SAMPLE SCHEME FOR THE REFERENCES
% ----------------------------------------------------------------------------

% ----------------------------------------------------------------------------

\end{footnotesize}
\end{document}